\documentclass[conference]{IEEEtran}

\usepackage[utf8]{inputenc}
\usepackage[T1]{fontenc}
\usepackage{cite}
\usepackage{amsmath,amssymb,amsfonts}
\usepackage{graphicx}
\usepackage{textcomp}
\usepackage{xcolor}
\usepackage{booktabs}
\usepackage{multirow}
\usepackage{url}
\usepackage{hyperref}
\usepackage{enumitem}

\hypersetup{
    colorlinks=true,
    linkcolor=black,
    citecolor=black,
    urlcolor=blue
}

\begin{document}

\title{Throughput Optimization as a Strategic Lever in Large-Scale AI Systems: Evidence from Dataloader and Memory Profiling Innovations}

\author{\IEEEauthorblockN{Mayank Jha}
\IEEEauthorblockA{Amazon Robotics\\
\texttt{mayankjh@amazon.com}}}

\maketitle

\begin{abstract}
The development of large-scale foundation models, particularly Large Language Models (LLMs), is constrained by significant computational and memory bottlenecks. These challenges elevate throughput optimization from a mere engineering task to a critical strategic lever, directly influencing training time, operational cost, and the feasible scale of next-generation models. This paper synthesizes evidence from recent academic and industry innovations to analyze key advancements in training efficiency. We examine architectural solutions to dataloader bottlenecks, such as the OVERLORD framework, which has demonstrated a 4.5$\times$ improvement in end-to-end training throughput. We investigate memory optimization techniques designed to overcome the GPU memory wall, including CPU offloading strategies like DeepSpeed's ZeRO-Offload, which enable the training of models far exceeding single-accelerator capacity. Furthermore, we explore the growing importance of compiler-centric optimizations, exemplified by Triton-distributed, which enables the joint optimization of computation, memory, and communication for substantial performance gains. The analysis is contextualized by advanced profiling tools and hardware characterization studies that identify and mitigate previously overlooked overheads like Dynamic Voltage and Frequency Scaling (DVFS). Findings indicate that a holistic, system-level approach, integrating innovations across data pipelines, memory management, network fabrics, and compiler technologies, is essential for accelerating AI development, managing costs, and pushing the boundaries of model scale.
\end{abstract}


\section{Introduction}

The field of artificial intelligence is increasingly dominated by large-scale foundation models, such as Google's Gemini and Meta's LLaMA series, which are trained on vast datasets and can be adapted to a wide array of downstream tasks~\cite{tiwari2025}. The training of these models represents a monumental engineering challenge, requiring dedicated supercomputer clusters with tens or even hundreds of thousands of accelerators. For instance, OpenAI's GPT-4 training utilized tens of thousands of GPUs on Microsoft Azure, while Meta is reportedly training its LLaMA~4 model on a cluster of over 100,000 NVIDIA H100 GPUs~\cite{tiwari2025}. This massive scale introduces profound bottlenecks related to data input, accelerator memory capacity, and inter-node communication, which collectively hinder performance and inflate operational costs, with the Total Cost of Ownership (TCO) for next-generation systems like the NVIDIA GB200 NVL72 being approximately 1.6 times higher than for H100 systems~\cite{patel2025}.

Consequently, throughput optimization has become a primary strategic lever for organizations developing these systems. Improvements in training efficiency directly translate into three critical business and research advantages: (1)~a reduction in time-to-train, which accelerates research and development cycles; (2)~minimization of operational costs, as the compute expenditure for a single training run can reach millions of dollars; and (3)~the ability to train larger, more capable models by overcoming fundamental hardware memory limitations. This paper aims to synthesize and analyze recent, high-impact innovations across the AI training stack---from data loading and memory management to compiler technologies and network optimization---that address these critical bottlenecks. By examining specific tools, architectural patterns, and research findings, this paper demonstrates how systematic throughput optimization is enabling the continued scaling and advancement of large-scale AI.

\begin{figure*}[t]
    \centering
    \includegraphics[width=0.95\textwidth]{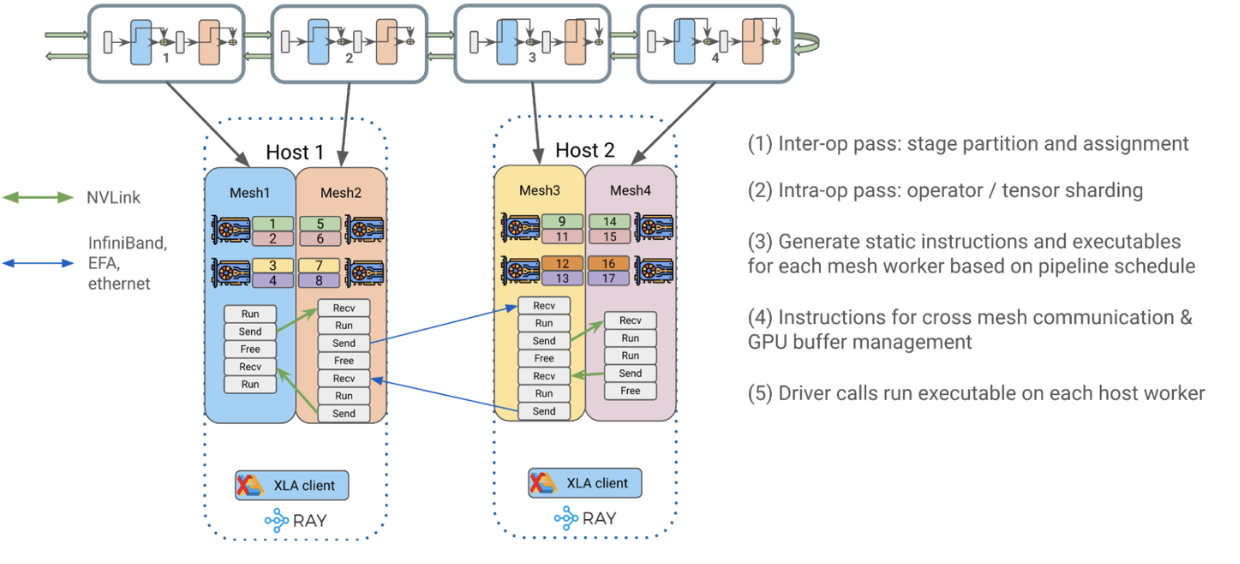}
    \caption{System-level execution model illustrating pipeline stage partitioning across hosts, intra-operator tensor sharding within each host, and cross-host communication over high-speed interconnects. Static execution schedules generate per-worker instructions that coordinate computation, communication, and GPU buffer management, exposing key sources of throughput and cost bottlenecks in large-scale training.}
    \label{fig:system_model}
\end{figure*}

\section{Literature Review}

The primary impediments to scaling AI training are well-documented hardware limitations, often referred to as ``walls.'' The most immediate is the \emph{GPU memory wall}, a significant bottleneck where the VRAM capacity of a single accelerator is insufficient to hold the model parameters, gradients, and optimizer states required for training~\cite{rajbhandari2020}. A common estimation for transformer-based models using an Adam optimizer is that the required VRAM is approximately 40 times the model's parameter count in billions~\cite{osc_gpu}. This means a relatively modest 7-billion parameter model requires at least 280\,GB of VRAM, far exceeding the capacity of a single state-of-the-art accelerator and mandating distributed training paradigms~\cite{osc_gpu}.

Table~\ref{tab:memory_wall} summarizes the estimated VRAM requirements for representative model sizes, illustrating the severity of the GPU memory wall.

\begin{table}[t]
\centering
\caption{Estimated VRAM Requirements for Training Transformer Models with Adam Optimizer ($\approx$40$\times$ Parameter Count)}
\label{tab:memory_wall}
\begin{tabular}{@{}lrr@{}}
\toprule
\textbf{Model Size} & \textbf{Est. VRAM} & \textbf{GPUs (80\,GB)} \\
\midrule
1B   & 40\,GB   & 1  \\
7B   & 280\,GB  & 4  \\
13B  & 520\,GB  & 7  \\
70B  & 2,800\,GB & 35 \\
175B & 7,000\,GB & 88 \\
\bottomrule
\end{tabular}
\end{table}

Beyond local memory, data movement itself is a projected bottleneck. As computational power (measured in FLOPs) grows, the ability of network and memory subsystems to feed the processing units becomes the limiting factor. Research projects that efficient scaling may become infeasible past $2 \times 10^{28}$ FLOPs for a three-month training run due to data movement constraints~\cite{erdil2024}. Even more fundamentally, a ``latency wall'' exists at approximately $2 \times 10^{31}$ FLOPs, where the time required for a single gradient step becomes shorter than the network latency for communication between nodes, a challenge that is particularly difficult to overcome as latency improves more slowly than other hardware metrics~\cite{erdil2024}. These theoretical and practical limits have spurred the development of a diverse ecosystem of software, hardware, and algorithmic solutions designed to manage memory hierarchies, optimize data pipelines, and efficiently orchestrate communication in large, distributed systems.

\section{Methodology}

This paper conducts a qualitative synthesis of contemporary research and technical literature focused on throughput optimization in large-scale AI training systems. The methodology involves a systematic review of peer-reviewed articles, technical reports from leading AI laboratories, and industry analyses published between 2024 and 2025. The selection criteria prioritized innovations with demonstrated, quantifiable performance improvements in real-world or large-scale simulation environments. The analysis is structured around key bottlenecks in the training pipeline: data loading, GPU memory management, distributed communication, and hardware utilization. By integrating findings from academic research (e.g., compiler design), open-source tools (e.g., DeepSpeed), and architectural best practices from major technology companies (e.g., Meta, NVIDIA), this study provides a holistic view of the strategic levers being employed to enhance training efficiency.

\section{Findings and Analysis}

This section synthesizes recent research and system-level evidence to analyze key throughput optimization strategies across data pipelines, memory management, distributed communication, and hardware-aware performance tuning in large-scale AI training systems.

\subsection{Innovations in Data Pipeline Management}

Data loading, or the process of feeding training samples from storage to accelerators, has emerged as a significant bottleneck, especially when using multi-source datasets and complex parallelism strategies. The OVERLORD architecture addresses this by disaggregating the data preprocessing pipeline into specialized actors: `Source Loaders' for sample-level transformations and `Data Constructors' for batch-level operations~\cite{zhao2025}. This design, coupled with a centralized data plane, eliminates redundant data access and reduces memory overhead. In production deployments on multi-thousand GPU clusters, OVERLORD achieved a 4.5$\times$ improvement in end-to-end training throughput and a 13.5$\times$ reduction in CPU memory usage, demonstrating that optimizing the data pipeline is a critical component of overall system performance~\cite{zhao2025}.

\subsection{Strategies for Overcoming the GPU Memory Wall}

To circumvent the physical memory limits of a single GPU, various offloading techniques have been developed. DeepSpeed's ZeRO-Offload is a prominent example, enabling models with over 10 billion parameters to be trained on a single GPU by delegating memory and computation to the host CPU and RAM~\cite{singh2022}. Its different stages allow for progressively offloading optimizer states, gradients, and even model parameters to system memory, freeing up valuable VRAM~\cite{singh2022}. A similar technique is CPU offloading of activations, where intermediate activation tensors are temporarily moved to CPU memory~\cite{sevegnani2025}. While effective, this introduces synchronization overhead that can impact GPU utilization. On specialized hardware like NVIDIA's Grace Hopper systems, Unified Memory provides a coherent memory space across CPU and GPU, though its efficiency is workload-dependent, showing minimal overhead for certain fine-tuning tasks but significant migration for others~\cite{sevegnani2025}. These strategies are not just optimizations but enabling technologies that allow for the training of models that would otherwise be impossible due to memory constraints.

Table~\ref{tab:memory_strategies} compares the key memory optimization strategies and their characteristics.

\begin{table}[t]
\centering
\caption{Comparison of GPU Memory Optimization Strategies}
\label{tab:memory_strategies}
\begin{tabular}{@{}p{1.6cm}p{2.2cm}p{1.8cm}p{1.6cm}@{}}
\toprule
\textbf{Strategy} & \textbf{Mechanism} & \textbf{Benefit} & \textbf{Trade-off} \\
\midrule
ZeRO Stage~1 & Offload optimizer states to CPU & Moderate VRAM savings & Low overhead \\
\addlinespace
ZeRO Stage~2 & Offload gradients + optimizer & High VRAM savings & Moderate overhead \\
\addlinespace
ZeRO Stage~3 & Offload all parameters to CPU & Train 10B+ on 1~GPU & Higher comm. cost \\
\addlinespace
Activation offload & Move activations to CPU RAM & Reduced peak VRAM & Sync overhead \\
\addlinespace
Unified Memory & Coherent CPU--GPU address space & Transparent migration & Hardware-dependent \\
\bottomrule
\end{tabular}
\end{table}

\subsection{Compiler-Centric Optimizations for Distributed Workloads}

Recent advancements have shifted optimization efforts to the compiler level, enabling more fine-grained control over hardware resources. Triton-distributed, an extension of the Triton compiler, is the first to support native overlapping optimizations for distributed AI workloads~\cite{zheng2025}. It integrates communication primitives compliant with the OpenSHMEM standard into a high-level Python programming model, allowing for the joint optimization of computation, memory access, and communication~\cite{zheng2025}. This approach has yielded speedups ranging from 1.09$\times$ to 44.97$\times$ over standard PyTorch with NCCL/RCCL baselines~\cite{zheng2025}. The framework's programming model follows the Multiple Programs Multiple Data (MPMD) paradigm, using concepts like `Symmetric Memory' and `Async-Task' to orchestrate parallel operations~\cite{zheng2025}.

\subsection{Network Fabric and Communication Optimization}

At scale, network performance is paramount. Meta found that tuning the standard DCQCN congestion control for RoCE (RDMA over Converged Ethernet) was challenging for training workloads, leading them to rely on Priority Flow Control (PFC) and a co-designed, receiver-driven traffic admission control mechanism~\cite{gangidi2024}. To improve load balancing, they also implemented Enhanced ECMP (E-ECMP) by hashing on the RoCE Queue Pair (QP) field, which improved AllReduce performance by up to 40\%~\cite{gangidi2024}. The choice of interconnect technology is also critical; benchmarks show that using NVIDIA Quantum-2 InfiniBand ($\sim$400\,GB/s) resulted in a 9$\times$ training speedup compared to standard 10\,Gbit Ethernet when finetuning a 12B parameter model, reducing the time per step from 39.8 to 4.4 seconds~\cite{maknee2025}.

Table~\ref{tab:interconnect} summarizes the impact of interconnect technology on distributed training performance.

\begin{table}[t]
\centering
\caption{Interconnect Impact on 12B Parameter Model Fine-tuning}
\label{tab:interconnect}
\begin{tabular}{@{}lccc@{}}
\toprule
\textbf{Interconnect} & \textbf{Bandwidth} & \textbf{Time/Step} & \textbf{Speedup} \\
\midrule
10\,Gbit Ethernet      & $\sim$1.25\,GB/s  & 39.8\,s & 1$\times$       \\
IB Quantum-2 (400G)    & $\sim$400\,GB/s   & 4.4\,s  & 9$\times$       \\
\bottomrule
\end{tabular}
\end{table}

\subsection{Advanced Profiling and Hardware-Specific Tuning}

Identifying the root causes of performance gaps requires sophisticated profiling tools. An analysis using the `Chopper' profiling tool on an AMD Instinct\texttrademark{} MI300X node found that frequency overhead from Dynamic Voltage and Frequency Scaling (DVFS) was the single largest contributor to the gap between theoretical and observed performance during Llama~3 8B training~\cite{kurzynski2025}. The same study revealed that the more deterministic memory allocation of FSDPv2 (a distributed training strategy) allowed GPUs to sustain approximately 20\% higher and more stable clock frequencies compared to FSDPv1, outweighing the overhead of extra data copy operations~\cite{kurzynski2025}. This highlights the importance of deep, hardware-aware profiling to uncover non-obvious optimization opportunities.

\subsection{Abstraction for Heterogeneous Hardware}

As AI infrastructure diversifies beyond a single vendor, hardware abstraction becomes crucial for portability and efficiency. The UniOrch framework addresses this by using a unified compilation system based on Triton and MLIR~\cite{wang_uniorch}. It captures PyTorch operator calls and converts them into a hardware-independent intermediate representation (Triton-IR). This IR is then optimized and lowered to hardware-specific backends like NVIDIA's PTX or Huawei's CANN IR, enabling a single codebase to run efficiently across different GPUs, NPUs, and DCUs~\cite{wang_uniorch}. This approach reduces development overhead and allows organizations to leverage a multi-vendor accelerator ecosystem.


Table~\ref{tab:summary} provides a consolidated summary of the key throughput optimization innovations analyzed in this paper.

\begin{table*}[t]
\centering
\caption{Summary of Key Throughput Optimization Innovations in Large-Scale AI Training}
\label{tab:summary}
\begin{tabular}{@{}llll@{}}
\toprule
\textbf{Optimization Domain} & \textbf{Innovation / Tool} & \textbf{Key Mechanism} & \textbf{Reported Improvement} \\
\midrule
Data Pipeline       & OVERLORD~\cite{zhao2025}            & Disaggregated actors, centralized data plane      & 4.5$\times$ throughput, 13.5$\times$ memory reduction \\
Memory Management   & ZeRO-Offload~\cite{singh2022}       & Progressive CPU/RAM offloading                    & Train 10B+ models on 1~GPU \\
Compiler            & Triton-distributed~\cite{zheng2025}  & MPMD with overlapping compute/comm                & 1.09$\times$--44.97$\times$ over PyTorch+NCCL \\
Network Fabric      & E-ECMP (Meta)~\cite{gangidi2024}     & QP-based hashing for load balancing               & Up to 40\% AllReduce improvement \\
Interconnect        & InfiniBand 400G~\cite{maknee2025}    & High-bandwidth RDMA interconnect                  & 9$\times$ training speedup vs. 10G Ethernet \\
Profiling           & Chopper~\cite{kurzynski2025}         & Multi-level GPU characterization                  & Identified DVFS as top perf. inhibitor \\
Hardware Abstraction& UniOrch~\cite{wang_uniorch}          & Triton/MLIR unified compilation                   & Single codebase across GPU/NPU/DCU \\
\bottomrule
\end{tabular}
\end{table*}

\section{Discussion}

The findings collectively illustrate a clear trend in large-scale AI training: optimization is shifting from isolated component-level improvements to a holistic, system-level co-design approach. The strategic importance of throughput is evident in how it directly addresses the primary objectives of reducing training time, minimizing operational costs, and enabling larger model scales. Innovations like OVERLORD demonstrate that the data pipeline, once a secondary consideration, can become a primary bottleneck and thus a fruitful area for optimization~\cite{zhao2025}. Similarly, memory offloading techniques like ZeRO-Offload are no longer just about cost savings; they are enabling technologies that fundamentally alter the limits of model size on existing hardware~\cite{singh2022}.

The rise of compiler-centric solutions like Triton-distributed signifies a crucial maturation in the field~\cite{zheng2025}. By providing developers with high-level abstractions to control low-level, hardware-specific operations---such as overlapping communication and computation---these tools democratize performance engineering and unlock gains that are difficult to achieve with library-based approaches alone. This is further supported by evidence from Meta's network engineering, where performance gains required co-design between the network fabric configuration (E-ECMP) and the collective communication library~\cite{gangidi2024}. The practical implications are significant: achieving state-of-the-art performance requires deep integration across hardware, networking, compiler, and application layers.

A key limitation of the current landscape is the high degree of specialization. Many advanced techniques are tailored to specific hardware interconnects, as seen in Triton-distributed's distinct ``swizzling'' optimizations for NVIDIA's NVSwitch versus AMD's full-mesh topology~\cite{zheng2025}. While this customization extracts maximum performance, it can increase software complexity and hinder portability. Frameworks like UniOrch aim to mitigate this through hardware abstraction layers, but a trade-off between performance and portability often remains~\cite{wang_uniorch}. Furthermore, while this paper focuses on throughput, other factors like model accuracy, convergence speed, and the energy efficiency of training are equally critical dimensions that require a more integrated analysis. The findings from profiling tools, which identified DVFS as a major performance inhibitor, suggest that there are still significant, non-obvious performance gains to be realized by closing the gap between theoretical hardware capabilities and real-world application performance~\cite{kurzynski2025}.

\section{Conclusion}

This paper has examined throughput optimization as a critical strategic lever in the era of large-scale AI. The evidence demonstrates that addressing bottlenecks across the entire training stack---from data ingestion with systems like OVERLORD, to memory constraints with techniques like ZeRO-Offload, and communication efficiency with compiler-level frameworks like Triton-distributed---is essential for progress. These innovations are not merely incremental improvements; they are fundamental enablers that reduce time-to-market, control escalating computational costs, and permit the exploration of larger, more powerful foundation models.

The analysis underscores that a holistic, system-aware approach is paramount, where hardware, software, and networking are co-designed for maximum efficiency. Future research should focus on further unifying these disparate optimizations, perhaps through more advanced compiler ecosystems that can automate hardware-specific tuning and abstract away the complexity of heterogeneous systems. Additionally, developing compiler-integrated, portable profiling tools that can automatically identify and mitigate subtle performance inhibitors like DVFS will be crucial for ensuring that the theoretical power of next-generation hardware is fully realized in practice.


\end{document}